# Opposition-Based Electromagnetism-Like for Global Optimization

*Erik Cuevas[1], +Diego Oliva, *Daniel Zaldivar, *Marco Pérez-Cisneros and +Gonzalo Pajares

*Departamento de Ciencias Computacionales
Universidad de Guadalajara, CUCEI
Av. Revolución 1500, Guadalajara, Jal, México
{[1]erik.cuevas, daniel.zaldivar, marco.perez}@cucei.udg.mx

+Dpto. Ingeniería del Software e Inteligencia Artificial,
Facultad Informática, Universidad Complutense,
Av. Complutense S/N, 28040, Madrid, Spain
doliva@estumail.ucm.es, pajares@fdi.ucm.es

**Abstract**

Electromagnetism-like Optimization (EMO) is a global optimization algorithm, particularly well-suited to solve problems featuring non-linear and multimodal cost functions. EMO employs searcher agents that emulate a population of charged particles which interact to each other according to electromagnetism's laws of attraction and repulsion. However, EMO usually requires a large number of iterations for a local search procedure; any reduction or cancelling over such number, critically perturb other issues such as convergence, exploration, population diversity and accuracy. This paper presents an enhanced EMO algorithm called OBEMO, which employs the Opposition-Based Learning (OBL) approach to accelerate the global convergence speed. OBL is a machine intelligence strategy which considers the current candidate solution and its opposite value at the same time, achieving a faster exploration of the search space. The proposed OBEMO method significantly reduces the required computational effort yet avoiding any detriment to the good search capabilities of the original EMO algorithm. Experiments are conducted over a comprehensive set of benchmark functions, showing that OBEMO obtains promising performance for most of the discussed test problems.

## 1. Introduction

Global Optimization (GO) [1,2] has issued applications for many areas of science [3], engineering [4], economics [5,6] and others whose definition requires mathematical modelling [7,8]. In general, GO aims to find the global optimum for an objective function which has been defined over a given search space. The difficulties associated with the use of mathematical methods over GO problems have contributed to the development of alternative solutions. Linear programming and dynamic programming techniques, for example, often have failed in solving (or reaching local optimum at) NP-hard problems which feature a large number of variables and non-linear objective functions. In order to overcome such problems, researchers have proposed metaheuristic-based algorithms for searching near-optimum solutions.

---
[1] Corresponding author, Tel +52 33 1378 5900, ext. 27714, E-mail: erik.cuevas@cucei.udg.mx

Metaheuristic algorithms are stochastic search methods that mimic the metaphor of biological or physical phenomena. The core of such methods lies on the analysis of collective behaviour of relatively simple agents working on decentralized systems. Such systems typically gather an agent's population that can communicate to each other while sharing a common environment. Despite a non-centralized control algorithm regulates the agent behaviour, the agent can solve complex tasks by analyzing a given global model and harvesting cooperation to other elements. Therefore, a novel global behaviour evolves from interaction among agents as it can be seen on typical examples that include ant colonies, animal herding, bird flocking, fish schooling, honey bees, bacteria, charged particles and many more. Some other metaheuristic optimization algorithms have been recently proposed to solve optimization problems, such as Genetic Algorithms (GA) [9], Particle Swarm Optimization (PSO) [10], Ant Colony Optimization (ACO) [11], Differential Evolution (DE) [12], Artificial Immune Systems (AIS) [13] and Artificial Bee Colony [14] and Gravitational Search Algorithm (GSA) [15].

Electromagnetism-like algorithm (EMO) is a relatively new population-based meta-heuristic algorithm which was firstly introduced by Birbil and Fang [16] to solve continuous optimization models using bounded variables. The algorithm imitates the attraction–repulsion mechanism between charged particles in an electromagnetic field. Each particle represents a solution and carries a certain amount of charge which is proportional to the solution quality (objective function). In turn, solutions are defined by position vectors which give real positions for particles within a multi-dimensional space. Moreover, objective function values of particles are calculated considering such position vectors. Each particle exerts repulsion or attraction forces over other population members; the resultant force acting over a particle is used to update its position. Clearly, the idea behind the EMO methodology is to move particles towards the optimum solution by exerting attraction or repulsion forces. Unlike other traditional meta-heuristics techniques such as GA, DE, ABC and AIS, whose population members exchange materials or information between each other, the EMO methodology assumes that each particle is influenced by all other particles in the population, mimicking other heuristics methods such as PSO and ACO. Although the EMO algorithm shares some characteristics with PSO and ACO, recent works have exhibited its better accuracy regarding optimal parameters [17-20], yet showing convergence [21]. EMO has been successfully applied to solve different sorts of engineering problems such as flow-shop scheduling [22], communications [23], vehicle routing [24], array pattern optimization in circuits [25], neural network training [26] control systems [27] and image processing [28].

EMO algorithm employs four main phases: initialization, local search, calculation and movement. The local search procedure is a stochastic search in several directions over all coordinates of each particle. EMO's main drawback is its computational complexity resulting from the large number of iterations which are commonly required during the searching process. The issue becomes worst as the dimension of the optimization problem increases. Several approaches, which simplify the local search, have been proposed in the literature to reduce EMO's computational effort. In [29] where Guan et al. proposed a discrete encoding for the particle set in order to reduce search directions at each dimension. In [30] and [31], authors include a new local search method which is based on a fixed search pattern and a shrinking strategy that aims to reduce the population size as the iterative process progresses. Additionally, in [17], a modified local search phase that employs the gradient descent method is adopted to enhance its computational complexity. Although all these approaches have improved the computational time which is required by the original EMO algorithm, recent works [27,32] have demonstrated that reducing or simplifying EMO's local search processes also affects other important properties, such as convergence, exploration, population diversity and accuracy.

On the other hand, the opposition-based learning (OBL), that has been initially proposed in [33], is a machine intelligence strategy which considers the current estimate and its correspondent opposite value (i.e., guess and opposite guess) at the same time to achieve a fast approximation for a current candidate solution. It has been mathematically proved [34-36] that an opposite candidate solution holds a higher probability for approaching the global optimum solution than a given random candidate, yet quicker. Recently, the concept of opposition has been used to accelerate metaheuristic-based algorithms such as GA [37], DE [38], PSO [39] and GSA [40].

In this paper, an Opposition-Based EMO called OBEMO is constructed by combining the opposition-based strategy and the standard EMO technique. The enhanced algorithm allows a significant reduction on the computational effort which required by the local search procedure yet avoiding any detriment to the good search capabilities and convergence speed of the original EMO algorithm. The proposed algorithm has been experimentally tested by means of a comprehensive set of complex benchmark functions. Comparisons to the original EMO and others state-of-the-art EMO-based algorithms [7] demonstrate that the OBEMO technique is faster for all test functions, yet delivering a higher accuracy. Conclusions on the conducted experiments are supported by statistical validation that properly supports the results.

The rest of the paper is organized as follows: Section 2 introduces the standard EMO algorithm. Section 3 gives a simple description of OBL and Section 4 explains the implementation of the proposed OBEMO algorithm. Section 5 presents a comparative study among OBEMO and other EMO variants over several benchmark problems. Finally, some conclusions are drawn in Section 6.

## 2. Electromagnetism - Like Optimization Algorithm (EMO)

EMO algorithm is a simple and direct search algorithm which has been inspired by the electro-magnetism phenomenon. It is based on a given population and the optimization of global multi-modal functions. In comparison to GA, it does not use crossover or mutation operators to explore feasible regions; instead it does implement a collective attraction–repulsion mechanism yielding a reduced computational cost with respect to memory allocation and execution time. Moreover, no gradient information is required as it employs a decimal system which clearly contrasts to GA. Few particles are required to reach converge as has been already demonstrated in [11].

EMO algorithm can effectively solve a special class of optimization problems with bounded variables in the form of:

$$\min f(x) \\ x \in [l, u] \quad , \tag{1}$$

where $[l,u] = \{x \in \Re^n \mid l_d \leq x_d \leq u_d,\ d = 1,2...n\}$ and $n$ being the dimension of the variable $x$, $[l,u] \subset \Re^n$, a nonempty subset and a real-value function $f:[l,u] \to \Re$. Hence, the following problem features are known:

- $n$: Dimensional size of the problem.
- $u_d$: The highest bound of the $k^{th}$ dimension.
- $l_d$: The lowest bound of the $k^{th}$ dimension.
- $f(x)$: The function to be minimized.

EMO algorithm has four phases [6]: initialization, local search, computation of the total force vector and movement. A deeper discussion about each stage follows.

**Initialization,** a number of $m$ particles is gathered as their highest ($u$) and lowest limit ($l$) are identified.

**Local search,** gathers local information for a given point $\mathbf{g}^p$, where $p \in (1,\ldots,m)$.

**Calculation of the total force vector,** charges and forces are calculated for every particle.

**Movement,** each particle is displaced accordingly, matching the corresponding force vector.

*2.1 Initialization*

First, the population of $m$ solutions is randomly produced at an initial state. Each $n$-dimensional solution is regarded as a charged particle holding a uniform distribution between the highest ($u$) and the lowest ($l$) limits. The optimum particle (solution) is thus defined by the objective function to be optimized. The procedure ends when all the $m$ samples are evaluated, choosing the sample (particle) that has gathered the best function value.

*2.2 Local Search*

The local search procedure is used to gather local information within the neighbourhood of a candidate solution. It allows obtaining a better exploration and population diversity for the algorithm.

Considering a pre-fixed number of iterations known as *ITER* and a feasible neighbourhood search $\delta$, the procedure iterates as follows: Point $\mathbf{g}^p$ is assigned to a temporary point $\mathbf{t}$ to store the initial information. Next, for a given coordinate $d$, a random number is selected ($\lambda_1$) and combined with $\delta$ as a step length, which in turn, moves the point $\mathbf{t}$ along the direction $d$, with a randomly determined sign ($\lambda_2$). If point $\mathbf{t}$ observes a better performance over the iteration number *ITER*, point $\mathbf{g}^p$ is replaced by $\mathbf{t}$ and the neighbourhood search for point $\mathbf{g}^p$ finishes, otherwise $\mathbf{g}^p$ is held. The pseudo-code is listed in Fig. 1.

In general, the local search for all particles can also reduce the risk of falling into a local solution but is time consuming. Nevertheless, recent works [17,32] have shown that eliminating, reducing or simplifying the local search process affects significantly the convergence, exploration, population diversity and accuracy of the EMO algorithm.

2.3 Total force vector computation

The total force vector computation is based on the *superposition principle* (Fig. 2) from the electro-magnetism theory which states: "the force exerted on a point via other points is inversely proportional to the distance between the points and directly proportional to the product of their charges" [41]. The particle moves following the resultant Coulomb's force which has been produced among particles as a charge-like value. In the EMO implementation, the charge for each particle is determined by its fitness value as follows:

$$q^p = \exp\left(-n \frac{f(\mathbf{g}^p) - f(\mathbf{g}^{best})}{\sum_{h=1}^{m}\left(f(\mathbf{g}^h) - f(\mathbf{g}^{best})\right)}\right), \forall p, \qquad (2)$$

where $n$ denotes the dimension of $\mathbf{g}^p$ and $m$ represents the population size. A higher dimensional problem usually requires a larger population. In Eq. (2), the particle showing the best fitness function value $\mathbf{g}^{best}$ is called the "*best particle*", getting the highest charge and attracting other particles holding high fitness values. The repulsion effect is applied to all other particles exhibiting lower fitness values. Both effects, attraction-repulsion are applied depending on the actual proximity between a given particle and the best-graded element.

| | | | |
|---|---|---|---|
| 1: | $Counter \leftarrow 1$ | 12: | $\mathbf{t}_d \leftarrow \mathbf{t}_d - \lambda_2(Length)$ |
| 2: | $Length \leftarrow \delta(\max\{u_d - l_d\})$ | 13: | **end if** |
| 3: | **for** $p=1$ **to** $m$ **do** | 14: | **if** $f(\mathbf{t}) < f(\mathbf{g}^p)$ **then** |
| 4: | **for** $d=1$ **to** $n$ **do** | 15: | $\mathbf{g}^p \leftarrow \mathbf{t}$ |
| 5: | $\lambda_1 \leftarrow U(0,1)$ | 16: | counter ← ITER − 1 |
| 6: | **while** $Counter < ITER$ **do** | 17: | **end if** |
| 7: | $\mathbf{t} \leftarrow \mathbf{g}^p$ | 18: | $Counter \leftarrow Counter + 1$ |
| 8: | $\lambda_2 \leftarrow U(0,1)$ | 19: | **end while** |
| 9: | **if** $\lambda_1 > 0.5$ **then** | 20: | **end for** |
| 10: | $\mathbf{t}_d \leftarrow \mathbf{t}_d + \lambda_2(Length)$ | 21: | **end for** |
| 11: | **Else** | 22: | $\mathbf{g}^{best} \leftarrow \arg\min\{f(\mathbf{g}^p), \forall p\}$ |

**Fig. 1.** Pseudo-code list for the local search algorithm

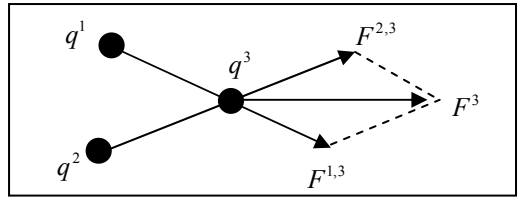

**Fig. 2.** The superposition principle

The overall resultant force between all particles determines the actual effect of the optimization process. The final force vector for each particle is evaluated under the Coulomb's law and the superposition principle as follows:

$$\mathbf{F}^p = \sum_{h \neq p}^{m} \begin{cases} (\mathbf{g}^h - \mathbf{g}^p)\dfrac{q^p q^h}{\|\mathbf{g}^h - \mathbf{g}^p\|^2} & if \quad f(\mathbf{g}^h) < f(\mathbf{g}^p) \\ (\mathbf{g}^p - \mathbf{g}^h)\dfrac{q^p q^h}{\|\mathbf{g}^h - \mathbf{g}^p\|^2} & if \quad f(\mathbf{g}^h) \geq f(\mathbf{g}^p) \end{cases}, \forall p \qquad (3)$$

where $f(\mathbf{g}^h) < f(\mathbf{g}^p)$ represents the attraction effect and $f(\mathbf{g}^h) \geq f(\mathbf{g}^p)$ represents the repulsion force (see Fig. 3). The resultant force of each particle is proportional to the product between charges and is inversely proportional to the distance between particles. In order to keep feasibility, the vector in expression (3) should be normalized as follows:

$$\hat{\mathbf{F}}^p = \frac{\mathbf{F}^p}{\|\mathbf{F}^p\|}, \quad \forall p. \tag{4}$$

*2.4. Movement*

The change of the *d*-coordinate for each particle *p* is computed with respect to the resultant force as follows:

$$g_d^p = \begin{cases} g_d^p + \lambda \cdot \hat{F}_d^p \cdot (u_d - g_d^p) & \text{if} \quad \hat{F}_d^p > 0 \\ g_d^p + \lambda \cdot \hat{F}_d^p \cdot (g_d^p - l_d) & \text{if} \quad \hat{F}_d^p \leq 0 \end{cases}, \forall p \neq best, \forall d \tag{5}$$

In Eq. (5), $\lambda$ is a random step length that is uniformly distributed between zero and one. $u_d$ and $l_d$ represent the upper and lower boundary for the *d*-coordinate, respectively. $\hat{F}_d^p$ represents the *d* element of $\hat{\mathbf{F}}^p$. If the resultant force is positive, then the particle moves towards the highest boundary by a random step length. Otherwise it moves toward the lowest boundary. The best particle does not move at all, because it holds the absolute attraction, pulling or repelling all others in the population.

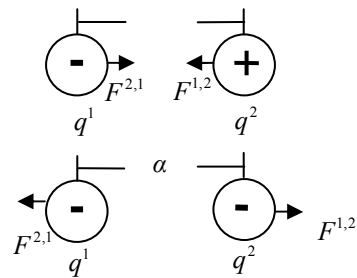

**Fig. 3.** Coulomb law: $\alpha$ represents the distance between charged particles, $q^1, q^2$ are the charges, and $F$ is the exerted force as has been generated by the charge interaction.

The process is halted when a maximum iteration number is reached or when the value $f(\mathbf{g}^{best})$ is near to zero or to the required optimal value.

**3. Opposition - based Learning (OBL).**

Opposition-based Learning [33] is a new concept in computational intelligence that has been employed to effectively enhance several soft computing algorithms [42,43]. The approach simultaneously evaluates a solution $x$ and its opposite solution $\bar{x}$ for a given problem, providing a renewed chance to find a candidate solution lying closer to the global optimum [34].

*3.1 Opposite number*

Let $x \in [l,u]$ be a real number, where $l$ and $u$ are the lowest and highest bound respectively. The opposite of $x$ is defined by:

$$\bar{x} = u + l - x \tag{6}$$

*3.2 Opposite point*

Similarly, the opposite number definition is generalized to higher dimensions as follows: Let $\mathbf{x} = (x_1, x_2, \ldots, x_n)$ be a point within a *n*-dimensional space, where $x_1, x_2, \ldots, x_n \in R$ and $x_i \in [l_i, u_i]$, $i \in 1, 2, \ldots, n$. The opposite point $\bar{\mathbf{x}} = (\bar{x}_1, \bar{x}_2, \ldots, \bar{x}_n)$ is defined by:

$$\bar{x}_i = u_i + l_i - x_i \tag{7}$$

*3.3 Opposition-based optimization*

Metaheuristic methods start by considering some initial solutions (initial population) and trying to improve them toward some optimal solution(s). The process of searching ends when some predefined criteria are satisfied. In the absence of a priori information about the solution, random guesses are usually considered. The computation time, among others algorithm characteristics, is related to the distance of these initial guesses taken from the optimal solution. The chance of starting with a closer (fitter) solution can be enhanced by simultaneously checking the opposite solution. By doing so, the fitter one (guess or opposite guess) can be chosen as an initial solution following the fact that, according to probability theory, 50% of the time a guess is further from the solution than its opposite guess [35]. Therefore, starting with the closer of the two guesses (as judged by their fitness values) has the potential to accelerate convergence. The same approach can be applied not only to initial solutions but also to each solution in the current population.

By applying the definition of an opposite point, the opposition-based optimization can be defined as follows: Let $\mathbf{x}$ be a point in a *n*-dimensional space (i.e. a candidate solution). Assume $f(\mathbf{x})$ is a fitness function which evaluates the quality of such candidate solution. According to the definition of opposite point, $\bar{\mathbf{x}}$ is the opposite of $\mathbf{x}$. If $f(\bar{\mathbf{x}})$ is better than $f(\mathbf{x})$, then $\mathbf{x}$ is updated with $\bar{\mathbf{x}}$, otherwise current point $\mathbf{x}$ is kept. Hence, the best point ($\mathbf{x}$ or $\bar{\mathbf{x}}$) is modified using known operators from the population-based algorithm.

Figure 4 shows the opposition-based optimization procedure. In the example, Fig. 4a and 4b represent the function to be optimized and its corresponding contour plot, respectively. By applying the OBL principles to the current population $P$ (see Fig. 4b), the three particles $\mathbf{x}_1$, $\mathbf{x}_2$ and $\mathbf{x}_3$ produce a new population $OP$, gathering particles $\bar{\mathbf{x}}_1$, $\bar{\mathbf{x}}_2$ and $\bar{\mathbf{x}}_3$. The three fittest particles from $P$ and $OP$ are selected as the new population $P'$. It can be seen from Fig. 4b that $\mathbf{x}_1$, $\bar{\mathbf{x}}_2$ and $\bar{\mathbf{x}}_3$ are three new members in $P'$. In this case, the transformation conducted on $\mathbf{x}_1$ did not provide a best chance of finding a candidate solution closer to the global optimum. Considering the OBL selection mechanism, $\bar{\mathbf{x}}_1$ is eliminated from the next generation.

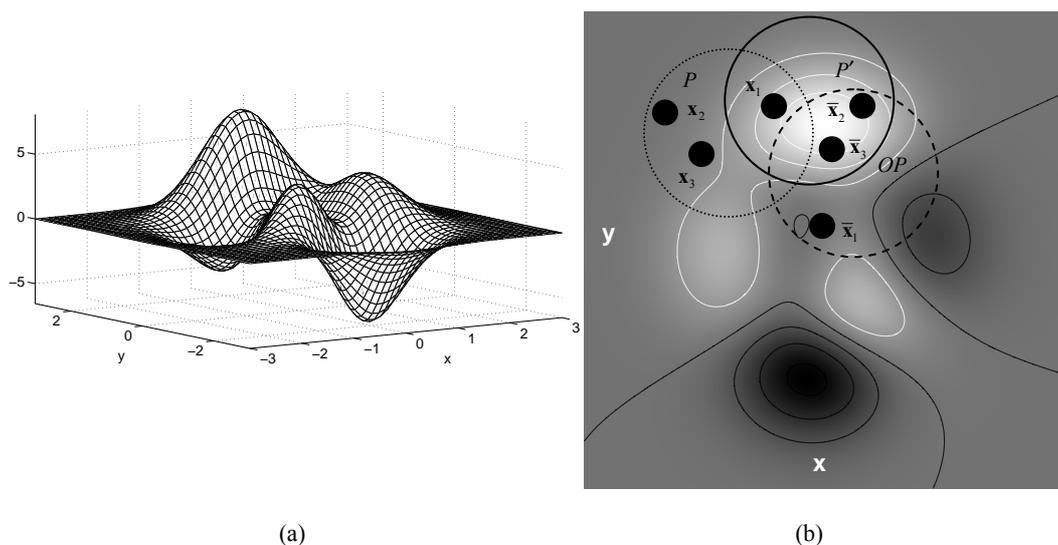

(a)          (b)
**Fig. 4.** The opposition-based optimization procedure: (a) Function to be optimized and (b) its contour plot. The current population $P$ includes particles $\mathbf{x}_1$, $\mathbf{x}_2$ and $\mathbf{x}_3$. The corresponding opposite population $OP$ is represented by $\bar{\mathbf{x}}_1$, $\bar{\mathbf{x}}_2$ and $\bar{\mathbf{x}}_3$. The final population $P'$ is obtained by the OBL selection mechanism yielding particles $\mathbf{x}_1$, $\bar{\mathbf{x}}_2$ and $\bar{\mathbf{x}}_3$.

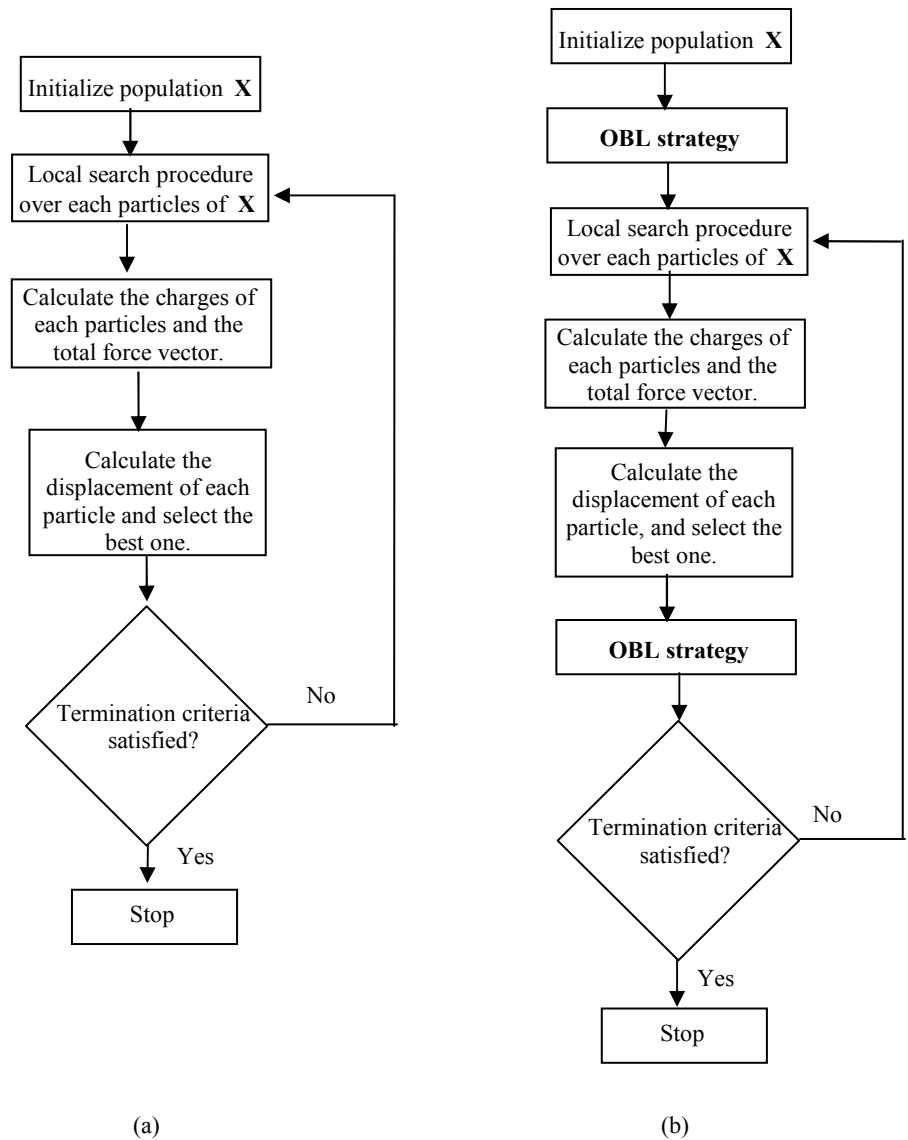

**Fig. 5.** Dataflow for: (a) the EMO method and (b) the OBEMO algorithm.

## 4. Opposition-based Electromagnetism-like Optimization (OBEMO)

Similarly to all metaheuristic-based optimization algorithms, two steps are fundamental for the EMO algorithm: the population initialization and the production of new generations by evolutionary operators. In the approach, the OBL scheme is incorporated to enhance both steps. However, the original EMO is considered as the main algorithm while the opposition procedures are embedded into EMO aiming to accelerate its convergence speed. Figure 5 shows a data flow comparison between the EMO and the OBEMO algorithm. The novel extended opposition procedures are explained in the following subsections.

### 4.1 Opposition-Based Population Initialization

In population-based meta-heuristic techniques, the random number generation is the common choice to create an initial population in absence of a priori knowledge. Therefore, as mentioned in Section 3, it is possible to obtain fitter starting candidate solutions by utilizing OBL despite no a-priori knowledge about the solution(s) is available. The following steps explain the overall procedure.

1) Initialize the population $\mathbf{X}$ with $N_P$ representing the number of particles.

2) Calculate the opposite population by

$$\bar{x}_i^j = u_i + l_i - x_i^j \quad (8)$$

$$i = 1, 2, \ldots, n; \quad j = 1, 2, \ldots, N_P$$

where $x_i^j$ and $\bar{x}_i^j$ denote the $i$th parameter of the $j$th particle of the population and its corresponding opposite particle.

3) Select the $N_P$ fittest elements from $\{\mathbf{X} \cup \bar{\mathbf{X}}\}$ as initial population.

### 4.2 Opposition-based production for new Generation

Starting from the current population, the OBL strategy can be used again to produce new populations. In this procedure, the opposite population is calculated and the fittest individuals are selected from the union of the current population and the opposite population. The following steps summarize the OBEMO implementation as follows:

**Step 1.** Generate $N_P$ initial random particles $\mathbf{x}^h$ to create the particle vector $\mathbf{X}$, with $h \in 1, 2, \ldots N_P$.

**Step 2.** Apply the OBL strategy by considering $N_P$ particles from vector $\mathbf{X}$ and generating the opposite vector $\overline{\mathbf{X}}$ through Eq. 7.

**Step 3.** Select the $N_P$ fittest particles from $\mathbf{X} \cup \overline{\mathbf{X}}$ according to $f(\cdot)$. These particles build the initial population $\mathbf{X}_0$.

**Step 4.** Calculate the local search procedure for each particle of $\mathbf{X}_0$ as follows: For a given dimension $d$, the particle $\mathbf{x}^h$ is assigned to a temporary point $y$ to store the initial information. Next, a random number is selected and combined with $\delta$ to yield the step length. Therefore, the point $y$ is moved along that direction. The sign is determined randomly. If $f(\mathbf{x}^h)$ is minimized, the particle $\mathbf{x}^h$ is replaced by $y$, ending the neighborhood-wide search for a particle $h$. The result is stored into the population vector $\mathbf{X}_{Local}$.

**Step 5.** Determine the best particle $\mathbf{x}^{best}$ of the population vector $\mathbf{X}_{Local}$ (with $\mathbf{x}^{best} \leftarrow \arg\min\{f(\mathbf{x}^h), \forall h\}$ ).

**Step 6.** Calculate the charge among particles using expression (2) and the vector force through Eq. (3). The particle showing the better objective function value holds a bigger charge and therefore a bigger attraction force.

**Step 7.** Change particle positions according to their force magnitude. The new particle's position is calculated by expression (5). $\mathbf{x}^{best}$ is not moved because it has the biggest force and attracts others particles to itself. The result is stored into the population vector $\mathbf{X}_{Mov}$.

**Step 8.** Apply the OBL strategy over the $m$ particles of the population vector $\mathbf{X}_{Mov}$, the opposite vector $\overline{\mathbf{X}}_{Mov}$ can be calculated through Eq. 7.

**Step 9.** Select the $m$ fittest particles from $\mathbf{X}_{Mov} \cup \overline{\mathbf{X}}_{Mov}$ according to $f(\cdot)$. Such particles represent the population $\mathbf{X}_0$.

**Step 10.** Increase the *Iteration* index. If *iteration = MAXITER or the* value of $f(X)$ is smaller than the pre-defined threshold value, then the algorithm is stopped and the flow jumps to step 11. Otherwise, it jumps to step 4.

**Step 11.** The best particle $\mathbf{x}^{best}$ is selected from the last iteration as it is considered as the solution.

## 5. Experimental results

In order to test the algorithm's performance, the proposed OBEMO is compared to the standard EMO and others state-of-the-art EMO-based algorithms. In this section, the experimental results are discussed in the following subsections:

(5.1) Test problems

(5.2) Parameter settings for the involved EMO algorithms

(5.3) Results and discussions

### 5.1. Test problems

A comprehensive set of benchmark problems, that includes 14 different global optimization tests, has been chosen for the experimental study. According to their use in the performance analysis, the functions are divided in two different sets: original test functions ($f_1 - f_9$) and multidimensional functions ($f_{10} - f_{14}$). Every function at this paper is considered as a minimization problem itself.

The original test functions, which are shown in Table 1, agree to the set of numerical benchmark functions presented by the original EMO paper at [16]. Considering that such function set is also employed by a vast majority of EMO-based new approaches, its use in our experimental study facilitates its comparison to similar works. More details can be found in [44].

The major challenge of an EMO-based approach is to avoid the computational complexity that arises from the large number of iterations which are required during the local search process. Since the computational complexity depends on the dimension of the optimization problem, one set of multidimensional functions (see Table 2) is used in order to assess the convergence and accuracy for each algorithm. Multidimensional functions include a set of five different functions whose dimension has been fixed to 30.

| Function | Search domain | Global minima |
|---|---|---|
| **Branin** $f_1(x_1,x_2) = (x_2 - \frac{5}{4\pi^2} x_1^2 + \frac{5}{\pi} x_1 - 6)^2 + 10(1 - \frac{1}{8\pi}) \cos x_1 + 10$ | $-5 \leq x_1 \leq 10$ $0 \leq x_2 \leq 15$ | 0.397887 |
| **Camel** $f_2(x_1,x_2) = -\frac{-x_1^2 + 4.5x_2^2 + 2}{e^{2x_2^2}}$ | $-2 \leq x_1, x_2 \leq 2$ | -1.031 |
| **Goldenstain-Price** | | |
| $f_3(x_1, x_2) = 1 + (x_1 + x_2 + 1)^2 \times (19 - 14x_1 + 13x_1^2 - 14x_2 + 6x_1x_2 + 3x_2^2)$ $\times (30 + 2x_1 - 3x_2)^2 \times (18 - 32x_1 + 12x_1^2 - 48x_2 - 36x_1x_2 + 27x_2^2)$ | $-2 \leq x_1, x_2 \leq 2$ | 3.0 |
| **Hartmann (3-dimensional)** $f_4(\mathbf{x}) = -\sum_{i=1}^{4} \alpha_i \exp\left[-\sum_{j=1}^{3} A_{ij}(x_j - P_{ij})^2\right]$ $\boldsymbol{\alpha} = [1, 1.2, 3, 3.2], \mathbf{A} = \begin{bmatrix} 3.0 & 10 & 30 \\ 0.1 & 10 & 35 \\ 3.0 & 10 & 35 \end{bmatrix}, \mathbf{P} = 10^{-4} \begin{bmatrix} 6890 & 1170 & 2673 \\ 4699 & 4387 & 7470 \\ 1091 & 8732 & 5547 \\ 381 & 5743 & 8828 \end{bmatrix}$ | $0 \leq x_i \leq 1$ $i = 1, 2, 3$ | -3.8627 |
| **Hartmann (6-dimensional)** $f_5(\mathbf{x}) = -\sum_{i=1}^{4} \alpha_i \exp\left[-\sum_{j=1}^{6} B_{ij}(x_j - Q_{ij})^2\right]$ $\boldsymbol{\alpha} = [1, 1.2, 3, 3.2], \mathbf{B} = \begin{bmatrix} 10 & 3 & 17 & 3.05 & 1.7 & 8 \\ 0.05 & 10 & 17 & 0.1 & 8 & 14 \\ 3 & 3.5 & 1.7 & 10 & 17 & 8 \\ 17 & 8 & 0.05 & 10 & 0.1 & 14 \end{bmatrix},$ $\mathbf{Q} = 10^{-4} \begin{bmatrix} 1312 & 1696 & 5569 & 124 & 8283 & 5886 \\ 2329 & 4135 & 8307 & 3736 & 1004 & 9991 \\ 2348 & 1451 & 3522 & 2883 & 3047 & 6650 \\ 4047 & 8828 & 8732 & 5743 & 1091 & 381 \end{bmatrix}$ | $0 \leq x_i \leq 1$ $i = 1, 2, 3, \ldots, 6$ | -3.8623 |
| **Shekel** $S_m$ **(4-dimensional)** $S_m(\mathbf{x}) = -\sum_{j=1}^{m} \left[\sum_{i=1}^{4}(x_i - C_{ij})^2 + \beta_j\right]^{-1}$ $\boldsymbol{\beta} = [1, 2, 2, 4, 4, 6, 3, 7, 5, 5]^T,$ $\mathbf{C} = \begin{bmatrix} 4.0 & 1.0 & 8.0 & 6.0 & 3.0 & 2.0 & 5.0 & 8.0 & 6.0 & 7.0 \\ 4.0 & 1.0 & 8.0 & 6.0 & 7.0 & 9.0 & 5.0 & 1.0 & 2.0 & 3.6 \\ 4.0 & 1.0 & 8.0 & 6.0 & 3.0 & 2.0 & 3.0 & 8.0 & 6.0 & 7.0 \\ 4.0 & 1.0 & 8.0 & 6.0 & 7.0 & 9.0 & 3.0 & 1.0 & 2.0 & 3.6 \end{bmatrix}$ | $0 \leq x_i \leq 1$ $i = 1, 2, 3, 4$ | |
| $f_6(\mathbf{x}) = S_5(\mathbf{x})$ | | -10.1532 |
| $f_7(\mathbf{x}) = S_7(\mathbf{x})$ | | -10.4029 |
| $f_8(\mathbf{x}) = S_{10}(\mathbf{x})$ | | -10.5364 |
| **Shubert** $f_9(x_1, x_2) = \left(\sum_{i=1}^{5} i \cos((i+1)x_1 + i)\right)\left(\sum_{i=1}^{5} i \cos((i+1)x_2 + i)\right)$ | $-10 \leq x_1, x_2 \leq 10$ | -186.73 |

**Table 1.** Optimization test functions corresponding to the original test set.

| Function | Search domain | Global minima |
|---|---|---|
| $f_{10}(\mathbf{x}) = \sum_{i=1}^{n}\left[x_i^2 - 10\cos(2\pi x_i) + 10\right]$ | $[-5.12, 5.12]^{30}$ | 0 |
| $f_{11}(\mathbf{x}) = -20\exp\left(-0.2\sqrt{\frac{1}{n}\sum_{i=1}^{n} x_i^2}\right) - \exp\left(\frac{1}{n}\sum_{i=1}^{n}\cos(2\pi x_i)\right) + 20$ | $[-32, 32]^{30}$ | 0 |
| $f_{12}(\mathbf{x}) = \frac{1}{4000}\sum_{i=1}^{n} x_i^2 - \prod_{i=1}^{n}\cos\left(\frac{x_i}{\sqrt{i}}\right) + 1$ | $[-600, 600]^{30}$ | 0 |
| $f_{13}(\mathbf{x}) = \frac{\pi}{n}\left\{10\sin(\pi y_1) + \sum_{i=1}^{n-1}(y_i - 1)^2\left[1 + 10\sin^2(\pi y_{i+1})\right] + (y_n - 1)^2\right\}$ $+ \sum_{i=1}^{n} u(x_i, 10, 100, 4)$ | $[-50, 50]^{30}$ | 0 |

$$y_i = 1 + \frac{x_i + 1}{4} \qquad u(x_i, a, k, m) = \begin{cases} k(x_i - a)^m & x_i > a \\ 0 & -a < x_i < a \\ k(-x_i - a)^m & x_i < -a \end{cases}$$

$$f_{14}(\mathbf{x}) = \sin^2(3\pi x_1) + \sum_{i=1}^{n}(x_i - 1)^2\left[1 + \sin^2(3\pi x_i + 1)\right]$$
$$+ (x_n - 1)^2\left[1 + \sin^2(2\pi x_n)\right] + \sum_{i=1}^{n} u(x_i, 5, 100, 4) \qquad [-50, 50]^{30} \qquad 0$$

**Table 2.** Multidimensional test function set.

*5.2. Parameter settings for the involved EMO algorithms*

The experimental set aims to compare four EMO-based algorithms including the proposed OBEMO. All algorithms face 14 benchmark problems. The algorithms are listed below:

- Standard EMO algorithm [16];
- Hybridizing EMO with descent search (HEMO) [17];
- EMO with fixed search pattern (FEMO) [30];
- The proposed approach OBEMO.

For the original EMO algorithm described in [16] and the proposed OBEMO, the parameter set is configured considering: $\delta = 0.001$ and *LISTER*=4. For the HEMO, the following experimental parameters are considered: $LsIt_{\max} = 10$, $\varepsilon_r = 0.001$ and $\gamma = 0.00001$. Such values can be assumed as the best configuration set according to [17]. Diverging from the standard EMO and the OBEMO algorithm, the HEMO method reduces the local search phase by only processing the best found particle $\mathbf{x}^{best}$. The parameter set for the FEMO approach is defined by considering the following values: $N_{fe}^{\max} = 100$, $N_{ls}^{\max} = 10$, $\delta = 0.001$, $\delta^{\min} = 1 \times 10^{-8}$ and $\varepsilon_\delta = 0.1$. All aforementioned EMO-based algorithms use the same population size of $m = 50$.

*5.3. Results and discussions*

Original test functions set

On this test set, the performance of the OBEMO algorithm is compared to standard EMO, HEMO and FEMO, considering the original test functions set. Such functions, presented in Table 1, hold different dimensions and one known global minimum. The performance is analyzed by considering 35 different executions for each algorithm. The case of no significant changes in the solution being registered (i.e. smaller than $10^{-4}$) is considered as stopping criterion.

The results, shown by Table 3, are evaluated assuming the averaged best value *f(x)* and the averaged number of executed iterations (*MAXITER*). Figure 6 shows the optimization process for the function $f_3$ and $f_6$. Such function values correspond to the best case for each approach that is obtained after 35 executions.

| | **Function** | $f_1$ | $f_2$ | $f_3$ | $f_4$ | $f_5$ | $f_6$ | $f_7$ | $f_8$ | $f_9$ |
|---|---|---|---|---|---|---|---|---|---|---|
| | Dimension | 2 | 2 | 2 | 3 | 6 | 4 | 4 | 4 | 2 |
| **EMO** | **Averaged best values *f(x)*** | 0.3980 | -1.015 | 3.0123 | -3.7156 | -3.6322 | -10.07 | -10.23 | -10.47 | -186.71 |
| | **Averaged MAXITER** | 103 | 128 | 197 | 1.59E+03 | 1.08E+03 | 30 | 31 | 29 | 44 |
| **OBEM** | **Averaged best values *f(x)*** | 0.3980 | -1.027 | 3.0130 | -3.7821 | -3.8121 | -10.11 | -10.22 | -10.50 | -186.65 |
| | **Averaged MAXITER** | 61 | 83 | 101 | 1.12E+03 | 826 | 18 | 19 | 17 | 21 |
| **HEMO** | **Averaged best values *f(x)*** | 0.5151 | -0.872 | 3.413 | -3.1187 | -3.0632 | -9.041 | -9.22 | -9.1068 | -184.31 |
| | **Averaged MAXITER** | 58 | 79 | 105 | 1.10E+03 | 805 | 17 | 18 | 15 | 22 |
| **FEMO** | **Averaged best values *f(x)*** | 0.4189 | -0.913 | 3.337 | -3.3995 | -3.2276 | -9.229 | -9.88 | -10.18 | -183.88 |
| | **Averaged MAXITER** | 63 | 88 | 98 | 1.11E+03 | 841 | 21 | 22 | 19 | 25 |

**Table 3.** Comparative results for the EMO, the OBEMO, the HEMO and the FEMO algorithms considering the original test functions set (Table 1).

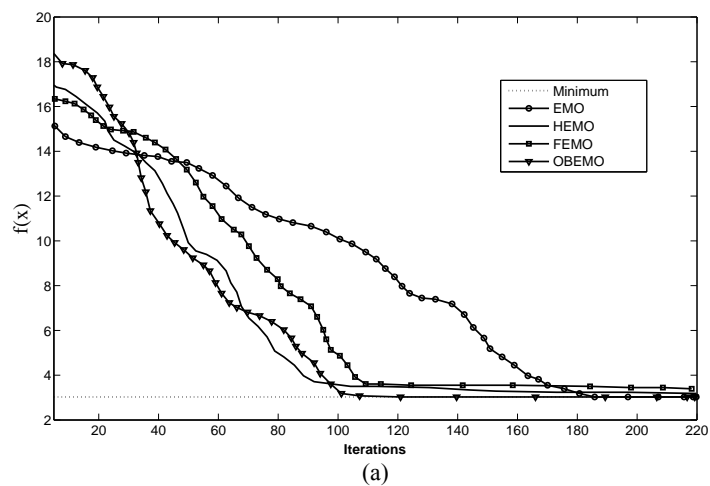

(a)

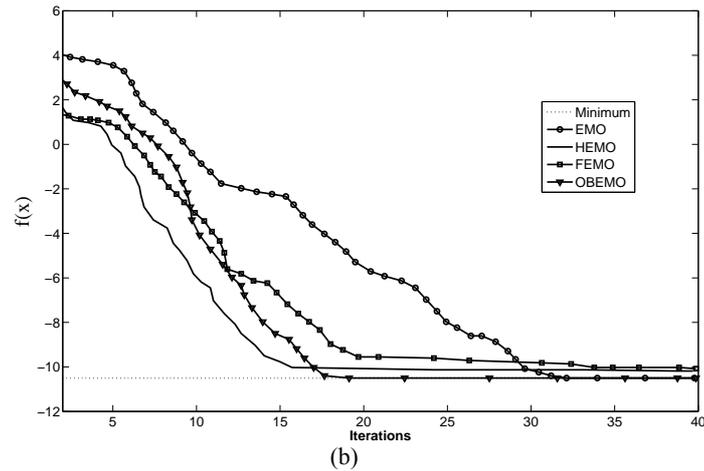

(b)

**Fig. 6.** Comparison of the optimization process for two original test functions: (a) $f_3$ and (b) $f_6$.

In order to statistically analyse the results in Table 3, a non-parametric significance proof known as the Wilcoxon's rank test [45-47] has been conducted. Such proof allows assessing result differences among two related methods. The analysis is performed considering a 5% significance level over the "averaged best value of *f(x)*" and the "averaged number of executed iterations of *MAXITER*" data. Table 4 and Table 5 reports the *p*-values produced by Wilcoxon's test for the pair-wise comparison of the "averaged best value" and the "averaged number of executed iterations" respectively, considering three groups. Such groups are formed by OBEMO vs. EMO, OBEMO vs. HEMO and OBEMO vs. FEMO. As a null hypothesis, it is assumed that there is no difference between the values of the two algorithms. The alternative hypothesis considers an actual difference between values from both approaches. The results obtained by the Wilcoxon test indicate that data cannot be assumed as occurring by coincidence (i.e. due to the normal noise contained in the process).

Table 4 considers the Wilcoxon analysis with respect to the "averaged best value" of *f(x)*. The *p*-values for the case of OBEMO vs EMO are larger than 0.05 (5% significance level) which is a strong evidence supporting the null hypothesis which indicates that there is no significant difference between both methods. On the other hand, in cases for the *p*-values corresponding to the OBEMO vs HEMO and OBEMO vs FEMO, they are less than 0.05 (5% significance level), which accounts for a significant difference between the "averaged best value" data among methods. Table 5 considers the Wilcoxon analysis with respect to the "averaged number of executed iterations" values. Applying the same criteria, it is evident that there is a significant difference between the OBEMO vs. EMO case, despite the OBEMO vs HEMO and OBEMO vs FEMO cases offering similar results.

| Function | *p*-Values | | |
|---|---|---|---|
| | OBEMO vs. EMO | OBEMO vs. HEMO | OBEMO vs. FEMO |
| $f_1$ | 0.3521 | 1.21E-04 | 1.02E-04 |
| $f_2$ | 0.4237 | 1.05E-04 | 0.88E-04 |
| $f_3$ | 0.2189 | 4.84E-05 | 3.12E-05 |
| $f_4$ | 0.4321 | 1.35E-05 | 1.09E-05 |
| $f_5$ | 0.5281 | 2.73E-04 | 2.21E-04 |
| $f_6$ | 0.4219 | 1.07E-04 | 0.77E-04 |
| $f_7$ | 0.3281 | 3.12E-05 | 2.45E-05 |
| $f_8$ | 0.4209 | 4.01E-05 | 3.62E-05 |
| $f_9$ | 0.2135 | 1.86E-05 | 1.29E-05 |

**Table 4.** Results from Wilcoxon's ranking test considering the "averaged best value of *f*(*x*)".

| Function | *p*-Values | | |
|---|---|---|---|
| | OBEMO vs. EMO | OBEMO vs. HEMO | OBEMO vs. FEMO |
| $f_1$ | 2.97E-04 | 0.2122 | 0.2877 |
| $f_2$ | 3.39E-04 | 0.1802 | 0.2298 |
| $f_3$ | 8.64E-09 | 0.1222 | 0.1567 |
| $f_4$ | 7.54E-05 | 0.2183 | 0.1988 |
| $f_5$ | 1.70E-04 | 0.3712 | 0.3319 |
| $f_6$ | 5.40E-13 | 0.4129 | 0.3831 |
| $f_7$ | 7.56E-04 | 0.3211 | 0.3565 |
| $f_8$ | 1.97E-04 | 0.2997 | 0.2586 |
| $f_9$ | 1.34E-05 | 0.3521 | 0.4011 |

**Table 5.** Results from Wilcoxon's ranking test considering the "averaged number of executed iterations".

Multidimensional functions

In contrast to the original functions, Multidimensional functions exhibit many local minima/maxima which are, in general, more difficult to optimize. In this section the performance of the OBEMO algorithm is compared to the EMO, the HEMO and the FEMO algorithms, considering functions in Table 2. This comparison reflects the algorithm's ability to escape from poor local optima and to locate a near-global optimum, consuming the least number of iterations. The dimension of such functions is set to 30. The results (Table 6) are averaged over 35 runs reporting the "averaged best value" and the "averaged number of executed iterations" as performance indexes.

|  | Function | $f_{10}$ | $f_{11}$ | $f_{12}$ | $f_{13}$ | $f_{14}$ |
|---|---|---|---|---|---|---|
|  | Dimension | 30 | 30 | 30 | 30 | 30 |
| EMO | Averaged best values $f(x)$ | 2.12E-05 | 1.21E-06 | 1.87E-05 | 1.97E-05 | 2.11E-06 |
| EMO | Averaged *MAXITER* | 622 | 789 | 754 | 802 | 833 |
| OBEM | Averaged best values $f(x)$ | 3.76E-05 | 5.88E-06 | 3.31E-05 | 4.63E-05 | 3.331E-06 |
| OBEM | Averaged *MAXITER* | 222 | 321 | 279 | 321 | 342 |
| HEMO | Averaged best values $f(x)$ | 2.47E-02 | 1.05E-02 | 2.77E-02 | 3.08E-02 | 1.88E-2 |
| HEMO | Averaged *MAXITER* | 210 | 309 | 263 | 307 | 328 |
| FEMO | Averaged best values $f(x)$ | 1.36E-02 | 2.62E-02 | 1.93E-02 | 2.75E-02 | 2.33E-02 |
| FEMO | Averaged *MAXITER* | 241 | 361 | 294 | 318 | 353 |

**Table 6.** Comparative results for the EMO, OBEMO, HEMO and the FEMO algorithms being applied to the multidimensional test functions (Table 2).

The Wilcoxon rank test results, presented in Table 7, shows that the p-values (regarding to the "averaged best value" values of Table 6) for the case of OBEMO vs EMO, indicating that there is no significant difference between both methods. p-values corresponding to the OBEMO vs HEMO and OBEMO vs FEMO show that there is a significant difference between the "averaged best" values among the methods. Figure 7 shows the optimization process for the function   and  . Such function values correspond to the best case, for each approach, obtained after 35 executions.

Table 8 considers the Wilcoxon analysis with respect to the "averaged number of executed iterations" values of Table 6. As it is observed, the outcome is similar to the results from last test on the original functions.

| Function | *p*-Values | | |
|---|---|---|---|
|  | OBEMO vs. EMO | OBEMO vs. HEMO | OBEMO vs. FEMO |
| $f_{10}$ | 0.2132 | 3.21E-05 | 3.14E-05 |
| $f_{11}$ | 0.3161 | 2.39E-05 | 2.77E-05 |
| $f_{12}$ | 0.4192 | 5.11E-05 | 1.23E-05 |
| $f_{13}$ | 0.3328 | 3.33E-05 | 3.21E-05 |
| $f_{14}$ | 0.4210 | 4.61E-05 | 1.88E-05 |

**Table 7.** Results from Wilcoxon's ranking test considering the "best averaged values".

| Function | *p*-Values | | |
|---|---|---|---|
|  | OBEMO vs. EMO | OBEMO vs. HEMO | OBEMO vs. FEMO |
| $f_{10}$ | 3.78E-05 | 0.1322 | 0.2356 |
| $f_{11}$ | 2.55E-05 | 0.2461 | 0.1492 |
| $f_{12}$ | 6.72E-05 | 0.3351 | 0.3147 |
| $f_{13}$ | 4.27E-05 | 0.2792 | 0.2735 |
| $f_{14}$ | 3.45E-05 | 0.3248 | 0.3811 |

**Table 8.** Results from Wilcoxon's ranking test considering the "averaged number of executed iterations"

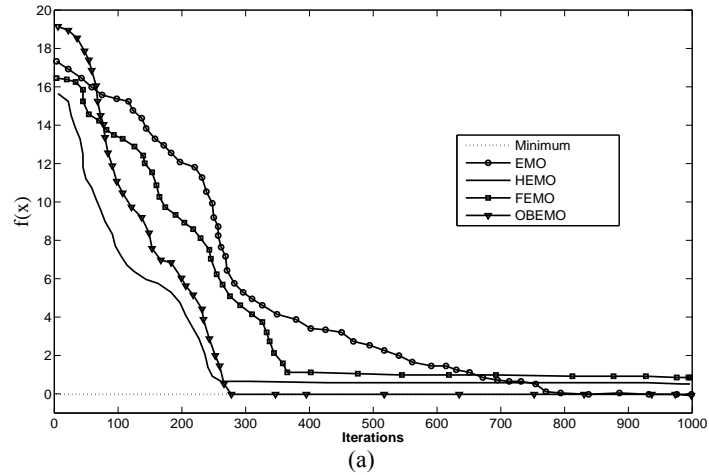

(a)

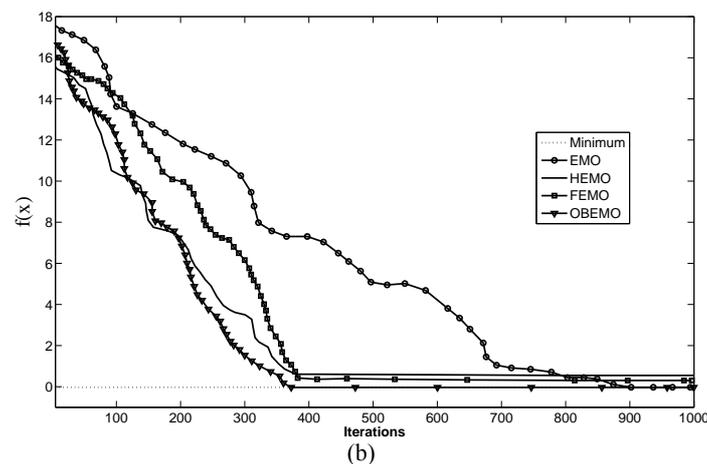

(b)

**Fig. 7.** Optimization process comparison for two multidimensional test functions: (a) $f_{12}$ and (b) $f_{14}$.

## 6. Conclusions

In this paper, an Opposition-Based EMO, named as OBEMO, has been proposed by combining the opposition-based learning (OBL) strategy and the standard EMO technique. The OBL is a machine intelligence strategy which considers, at the same time, a current estimate and its opposite value to achieve a fast approximation for a given candidate solution. The standard EMO is enhanced by using two OBL steps: the population initialization and the production of new generations. The enhanced algorithm significantly reduces the required computational effort yet avoiding any detriment to the good search capabilities of the original EMO algorithm.

A set of 14 benchmark test functions has been employed for experimental study. Results are supported by a statistically significant framework (Wilcoxon test [45-47]) to demonstrate that the OBEMO is as accurate as the standard EMO yet requiring a shorter number of iterations. Likewise, it is as fast as others state-of-the-art EMO-based algorithms such as HEMO [7] and FEMO [30], still keeping the original accuracy.

Although the results offer evidence to demonstrate that the Opposition-Based EMO method can yield good results on complicated optimization problems, the paper's aim is not to devise an optimization algorithm that could beat all others currently available, but to show that the Opposition-based Electromagnetism-like method can effectively be considered as an attractive alternative for solving global optimization problems.

**References**

[1] Songbo Tan, Xueqi Cheng and Hongbo Xu. An efficient global optimization approach for rough set based dimensionality reduction, International Journal of Innovative Computing, Information and Control, 3(3), 2007, 725-736.

[2] Ali Borji and Mandana Hamidi. A new approach to global optimization motivated by parliamentary political competitions. International Journal of Innovative Computing, Information and Control, 5(6), 2009, 1643-1653.

[3] Chia-Ning Yang, Kuo-Si Huang, Chang-Biau Yang and Chie-Yao Hsu. Error-tolerant minimum finding with DNA computing. International Journal of Innovative Computing, Information and Control, 5(10(A)), 2009, 3045-3057.

[4] Weijun Gao and Hongbo Ren. An optimization model based decision support system for distributed energy systems planning. International Journal of Innovative Computing, Information and Control, 7(5(B)), 2011, 2651-2668.

[5] Chunhui Xu, Jie Wang and Naoki Shiba. Multistage portfolio optimization with var as risk measure. International Journal of Innovative Computing, Information and Control, 3(3), 2007, 709-724.

[6] Jui-Fang Chang. A performance comparison between genetic algorithms and particle swarm optimization applied in constructing equity portfolios. International Journal of Innovative Computing, Information and Control, 5(12(B)), 2009, 5069-5079.

[7] Yoshiki Takeuchi. Optimization of linear observations for the stationary kalman filter based on a generalized water filling theorem. International Journal of Innovative Computing, Information and Control, 4(1), 2008, 211-230.

[8] Akbar H. Borzabadi, Mohammad E. Sadjadi and Behzad Moshiri. A numerical scheme for approximate optimal control of nonlinear hybrid systems. International Journal of Innovative Computing, Information and Control, 6(6), 2010, 2715-2724.

[9] Holland, J.H., Adaptation in Natural and Artificial Systems, University of Michigan Press, Ann Arbor, MI, 1975.

[10] J. Kennedy, R. Eberhart, Particle swarm optimization, in: IEEE International Conference on Neural Networks (Piscataway, NJ), 1995, pp. 1942–1948.

[11] M. Dorigo, V. Maniezzo, A. Colorni, Positive feedback as a search strategy, Technical Report 91-016, Politecnico di Milano, Italy, 1991.

[12] K. Price, R. Storn, A. Lampinen, Differential Evolution a Practical Approach to Global Optimization, Springer Natural Computing Series, 2005.

[13] Colin Fyfe and Lakhmi Jain. Teams of intelligent agents which learn using artificial immune systems.  Journal of Network and Computer Applications, 29(2-3), 2005, 147-159.

[14] D. Karaboga. An idea based on honey bee swarm for numerical optimization, technical report-TR06,Erciyes University, Engineering Faculty, Computer Engineering Department 2005.

[15] E. Rashedia, H. Nezamabadi-pour, S. Saryazdi. Filter modeling using Gravitational Search Algorithm. Engineering Applications of Artificial Intelligence, 24(1), 2011, 117-122.

[16] S. İlker Birbil and Shu-Cherng Fang. An Electromagnetism-like Mechanism for Global Optimization. Journal of Global Optimization, Vol 25: 263–282, (2003).

[17] A. Rocha, E. Fernandes, Hybridizing the electromagnetism-like algorithm with descent search for solving engineering design problems. International Journal of Computer Mathematics, 86 (2009) 1932–1946.

[18] A. Rocha, E. Fernandes, Modified movement force vector in an electromagnetism-like mechanism for global optimization. Optimization Methods & Software 24 (2009)  253–270.

[19] C.S. Tsou, C.H. Kao, Multi-objective inventory control using electromagnetism-like metaheuristic. International Journal of Production Research, 46 (2008) 3859–3874.

[20] P. Wu, Y. Wen-Hung, W. Nai-Chieh, An electromagnetism algorithm of neural network analysis an application to textile retail operation. Journal of the Chinese Institute of Industrial Engineers, 21 (2004) 59 – 67.

[21] Birbil, S. I., S. C. Fang, and R. L. Sheu, On the convergence of a population-based global optimization algorithm. Journal of Global Optimization, Vol. 30, No. 2, 301–318, 2004.

[22] B. Naderi, R. Tavakkoli-Moghaddam, M. Khalili, Electromagnetism-like mechanism and simulated annealing algorithms for flowshop scheduling problems minimizing the total weighted tardiness and makespan, Knowledge-Based Systems, 23 (2010)  77-85.

[23] Ho-Lung Hung and Yung-Fa Huang. Peak to average power ratio reduction of multicarrier transmission systems using electromagnetism-like method. International Journal of Innovative Computing, Information and Control, 7(5(A)), 2011, 2037-2050.

[24] A. Yurtkuran, E. Emel, A new Hybrid Electromagnetism-like Algorithm for capacitated vehicle routing problems. Expert Systems with Applications, 37 (2010) 3427-3433.

[25] J. Jhen-Yan, L. Kun-Chou, Array pattern optimization using electromagnetism-like algorithm. AEU - International Journal of Electronics and Communications, 63 (2009) 491-496.

[26] P. Wu, Y. Wen-Hung, W. Nai-Chieh, An electromagnetism algorithm of neural network analysis an application to textile retail operation. Journal of the Chinese Institute of Industrial Engineers, 21 (2004) 59 – 67.

[27] C.H. Lee, F.K. Chang, Fractional-order PID controller optimization via improved electromagnetism-like algorithm, Expert Systems with Applications, 37 (2010) 8871-8878.

[28] E. Cuevas, D. Oliva, D. Zaldivar, M. Pérez-Cisneros, H. Sossa. Circle detection using electro-magnetism optimization. Information Sciences. 182(1), (2012), 40-55.

[29] Xianping Guan, Xianzhong Dai and Jun Li. Revised electromagnetism-like mechanism for flow path design of unidirectional AGV systems. International Journal of Production Research 49(2), (2011), 401–429.

[30] Ana Maria A.C. Rocha and Edite Fernandes. Numerical Experiments with a Population Shrinking Strategy within a Electromagnetism-like Algorithm. Journal of Mathematics and Computers in Simulation. 1(3), 2007, 238-243.

[31] Ana Maria A.C. Rocha and Edite Fernandes. Numerical study of augmented Lagrangian algorithms for constrained global optimization. Optimization, 60(10-11), 2011, 1359-1378.

[32] Ching-Hung Lee, Fu-Kai Chang, Che-Ting Kuo, Hao-Hang Chang. A hybrid of electromagnetism-like mechanism and back-propagation algorithms for recurrent neural fuzzy systems design. International Journal of Systems Science, 43(2), 2012, 231-247.

[33] Tizhoosh H.R. Opposition-based learning: a new scheme for machine intelligence. In: Proceedings of international conference on computational intelligence for modeling control and automation, 2005, pp 695–701.

[34] S.Rahnamayn, H.R.Tizhoosh, M.Salama. A Novel Population Initialization Method for Accelerating Evolutionary Algorithms, Computers and Mathematics with Applications, Volume 53 , Issue 10 , pp.1605-1614, 2007.

[35] Rahnamayan S, Tizhoosh HR, Salama MMA. Opposition versus randomness in soft computing techniques. Elsevier J Appl. Soft Comput 8, 2008, 906–918.

[36] Hui Wang , Zhijian Wu, Shahryar Rahnamayan. Enhanced opposition-based differential evolution for solving high-dimensional continuous optimization problems. Soft Comput. 2010, DOI 10.1007/s00500-010-0642-7.

[37] Muhammad Amjad Iqbal, Naveed Kazim Khan, Hasan Multaba and A. Rauf Baig. A novel function optimization approach using opposition based genetic algorithm with gene excitation. International Journal of Innovative Computing, Information and Control, 7(7(B)), 2011, 4263-4276.

[38] Rahnamayan S, Tizhoosh HR, Salama MMA. Opposition-based differential evolution. IEEE Trans Evol Comput 12(1), 2008, 64–79.

[39] Hui Wanga, Zhijian Wua, Shahryar Rahnamayan, Yong Liu, Mario Ventresca. Enhancing particle swarm optimization using generalized opposition-based learning. Information Sciences 181, (2011), 4699–4714.

[40] Binod Shaw, V. Mukherjee, S.P. Ghoshal. A novel opposition-based gravitational search algorithm for combined economic and emission dispatch problems of power systems. Electrical Power and Energy Systems 35, (2012), 21–33.

[41] Cowan, E. W.  Basic Electromagnetism, Academic Press, New York, 1968.

[42] Tizhoosh HR. Opposition-based reinforcement learning. J Adv Comput Intell Intell Inform 2006;10(3):578–85.

[43] Shokri M, Tizhoosh HR, Kamel M. Opposition-based Q(k) algorithm. In: Proc IEEE world congr comput intell; 2006. p. 646–53.

[44] Dixon, L. C.W. and G. P Szegö. The global optimization problem: An introduction. Towards Global Optimization 2, North-Holland, Amsterdam (1978), pp. 1–15.

[45] Wilcoxon F (1945) Individual comparisons by ranking methods. Biometrics 1:80–83.

[46] Garcia S, Molina D, Lozano M, Herrera F (2008) A study on the use of non-parametric tests for analyzing the evolutionary algorithms' behaviour: a case study on the CEC'2005 Special session on real parameter optimization. J Heurist. doi:10.1007/s10732-008-9080-4.

[47] J. Santamaría, O. Cordón, S. Damas, J.M. García-Torres, A. Quirin, Performance Evaluation of Memetic Approaches in 3D Reconstruction of Forensic Objects. Soft Computing, DOI: 10.1007/s00500-008-0351-7, in press (2008).